\documentclass[letterpaper, 10 pt, conference]{ieeeconf}  % Comment this line out if you need a4paper

\IEEEoverridecommandlockouts                              % This command is only needed if 
\usepackage{hyperref}
\usepackage{url}

\usepackage[utf8]{inputenc} % allow utf-8 input
\usepackage[T1]{fontenc}    % use 8-bit T1 fonts
\usepackage{hyperref}       % hyperlinks
\usepackage{url}            % simple URL typesetting
\usepackage{booktabs}       % professional-quality tables
\usepackage{amsfonts}       % blackboard math symbols
\usepackage{nicefrac}       % compact symbols for 1/2, etc.
\usepackage{microtype}      % microtypography
\usepackage{balance}
\usepackage{amsmath}
\usepackage{amssymb}
\usepackage{mathrsfs}
\usepackage{subfigure}
\usepackage{graphicx}
\usepackage[style=base]{caption}
\usepackage{tabularx}
\usepackage{algorithm} 
\usepackage{algorithmic}  
\usepackage[algo2e]{algorithm2e} 

\usepackage[nolist]{acronym}
\usepackage{bbm}
\usepackage{xcolor}
\usepackage{wrapfig}
\usepackage{float}
\usepackage{siunitx}
\usepackage{multirow}

\let\labelindent\relax
\usepackage{enumitem}

\newcommand{\p}{\mathbf{p}}

\newcommand{\cmt}[1]{}

\long\def\ignorethis#1{}

\newcommand{\etal}{{\em{et~al.}\ }}

\newcommand{\vc}[1]{\ensuremath{\mathbf{#1}}}

\newcommand{\pctab}{\hspace{0.2in}}

\title{\LARGE \bf
Policy Transfer via Kinematic Domain Randomization and Adaptation
}

\author{Ioannis Exarchos$^{1}$, Yifeng Jiang$^{1}$, Wenhao Yu$^{2}$, and C. Karen Liu$^{1}$% <-this % stops a space
\thanks{$^{1}$Department of Computer Science,
        Stanford University, Stanford, CA 94305
        {\tt\small \{exarchos\},\{yifengj\}@stanford.edu, karenliu@cs.stanford.edu}}%
\thanks{$^{2}$Robotics at Google, Mountain View, CA
        {\tt\small magicmelon@google.com}}%
}

\begin{document}

\maketitle
\thispagestyle{empty}
\pagestyle{empty}

%%%%%%%%%%%%%%%%%%%%%%%%%%%%%%%%%%%%%%%%%%%%%%%%%%%%%%%%%%%%%%%%%%%%%%%%%%%%%%%%
\begin{abstract}
Transferring reinforcement learning policies trained in physics simulation to the real hardware remains a challenge, known as the ``sim-to-real'' gap. Domain randomization is a simple yet effective technique to address dynamics discrepancies across source and target domains, but its success generally depends on heuristics and trial-and-error. In this work we investigate the impact of randomized parameter selection on policy transferability across different types of domain discrepancies. Contrary to common practice in which kinematic parameters are carefully measured while dynamic parameters are randomized, we found that virtually randomizing kinematic parameters (e.g., link lengths) during training in simulation generally outperforms dynamic randomization. Based on this finding, we introduce a new domain adaptation algorithm that utilizes simulated kinematic parameters variation. Our algorithm, Multi-Policy Bayesian Optimization, trains an ensemble of universal policies conditioned on virtual kinematic parameters and efficiently adapts to the target environment using a limited number of target domain rollouts. We showcase our findings on a simulated quadruped robot in five different target environments covering different aspects of domain discrepancies.

\end{abstract}

%%%%%%%%%%%%%%%%%%%%%%%%%%%%%%%%%%%%%%%%%%%%%%%%%%%%%%%%%%%%%%%%%%%%%%%%%%%%%%%%
\section{Introduction}

The advent of Deep Reinforcement Learning (DRL) has demonstrated a promising approach to design robotic controllers for diverse robot motor skills \cite{akkaya2019solving, yu2018siggraph, peng2018deepmimic, hwangbo2019learning}. Nevertheless, DRL comes with the caveat of a very high demand in training data; this constitutes direct training in the real world cumbersome, if not infeasible. High-fidelity computer simulators of physics offer a way to bypass this practical obstacle but introduce a further complication: policies trained in simulation often fail to transfer to the real world due to modeling discrepancies between simulated and real environments. This is known as \textit{sim-to-real gap} \cite{neunert2017off}. % yifeng: made it shorter

Domain randomization is an approach that directly addresses the sim-to-real gap issue. Numerous evidence has shown that randomizing the parameters in the simulator during policy training leads to a policy that can perform well when tested in a different, target environment \cite{TanRSS18, hwangbo2019learning, peng2018sim, tobin2017domain}. However, the process of domain randomization depends heavily on heuristics and trial-and-error. Conventionally, practitioners randomize a selected set of parameters that are believed to have high impact or are difficult to measure precisely in the target environment, such as center of mass, latency, or friction coefficients.

% Effective selection of randomized parameters indeed depends on the robot, the environment, and the task at hand, but having some guideline on randomized parameter selection for a specific family of tasks could be beneficial to researchers and practitioners in robotics. 
What would be the effect if we intentionally randomize simulation parameters that are known, or can be easily measured? In this work, we first systematically investigate the impact of randomized parameter selection for a quadruped locomotion task on a variety of reality gaps. We analyze the parameters of a dynamic system in two categories: \emph{Kinematic} and \emph{Dynamic} parameters. We define as kinematic parameters those parameters that are required for computing forward kinematics on a robot, such as the link lengths, joint orientation, or joint degrees-of-freedom, while dynamic parameters include everything else. We engineer three types of domain discrepancies that mimic reality gaps to analyze the impact of different parameter categories with finer granularity: Kinematic Gap, Dynamic Gap, and Environment Gap. Using the same definition, a kinematic gap refers to deviations in one or more kinematic parameters between the source and target environments. We add environment gap as a third case to indicate differences that impact the agent’s policy but not the parameters of the agent themselves. In the context of locomotion, the environment gap is typically related to the type of surface the robot is walking on.

% in the environments that are completely unmodelled by either kinematic or dynamics randomization.
% \yifeng{not sure if we want to change to unmodelled dynamics gap. collision are indeed randomized in dynamics DR so people may get confused. Unmodelled dynamics gap is of high importance in real world I think.} \karen{agree} \wenhao{How about mentioning explicitly that all the gaps we introduce here are unmodelled dynamics? Also, I changed the last sentence to a more vague phrase (type of terrain) instead of the very specific ones (collision and contact). Would that help avoid confusion?}

Contrary to common practice in which kinematic parameters are carefully measured while dynamic parameters are randomized, we found that randomizing kinematic parameters during training in simulation produces the best performing policy, across \textit{all} three type of gaps. On the other hand, randomizing dynamic parameters only produces a policy capable of overcoming the dynamic gap itself, but typically fails on other unmodelled gaps. %We hypothesize that randomizing kinematic parameters results in wider exploration of the state space because it has a more global impact on the system dynamics--it effectively modifies the Jacobian matrix of center of mass for each link in an articulated robot system. In contrast (and perhaps counter-intuitively), randomizing dynamic parameters affects only specific aspects of the dynamic system. For example, inertial properties only take effect when the joint acceleration is high, and friction coefficients only matter when particular contact states occur.
Based on the above observation, a natural next step is to investigate whether domain adaptation methods based on universal policies (UP) \cite{yu2018policy} can also work well when conditioning on varying kinematic parameters. To train such a UP, we randomize the training environment with varying robot geometries and inform the UP of the kinematics parameters as part of its policy input. This creates diverse control behaviors parameterized by kinematics values. During deployment in the target environment, despite the physical robot having a fixed geometry, we may still treat the virtual kinematics input as a control knob and search for the parameter values that maximize target domain task performance with a small amount of trials. UPs conditioned on kinematic parameters improve target domain performance as expected, but we further notice that they highly depend on the random seeds used to initialize the policy optimization -- some seeds are better for different types of gaps than others. As such, we introduce a simple algorithm, which we call Multi-Policy Bayesian Optimization (MPBO), that utilizes an ensemble of UPs from different random seeds to further improve domain generalization. MPBO combines Bayesian optimization (BO) \cite{mockus2012bayesian} and the Upper-Confidence-Bound Action Selection (UCBAS) algorithm \cite{sutton2018reinforcement} in multi-armed bandit problems to determine the most effective UP as well as its optimal kinematic parameters using a limited number of rollouts.

We show experiments in simulations using a Laikago quadruped robot model and five different target environments. Our results suggest that randomizing kinematic parameters leads to a more generalizable policy across multiple types of sim-to-real gaps. %, including those not directly related to the parameters being randomized. 
Additionally, we show that MPBO can further improve domain generalization for all the environments we evaluated, at only a moderate amount of rollouts in the target environment. The code is available at \url{https://github.com/iexarchos/PolicyTransferKinDRA.git}

\section{Related Work}
Despite the tremendous progress in Deep Reinforcement Learning (DRL) for training complex motor skills such as walking \cite{yu2018siggraph}, flying \cite{won2017train}, and parkour \cite{peng2018deepmimic}, applying DRL to real robotic control problems is still a challenging problem due to the sim-to-real gap. Efforts in  enabling simulation-trained policies to be applied to real robots have largely focused on two complementary fronts: 1) improving the simulation model to better match the real-world dynamics, and 2) improving the policy training process such that it can generalize to a large variety of situations including the real-world. To improve the simulation model, researchers have proposed new algorithms for making the simulation models more expressive, identified key factors that cause the reality gap and demonstrated successful deployment of simulation-trained policies on real robots \cite{hwangbo2019learning, TanRSS18, hanna2017grounded}. For instance, Hwangbo \etal trained a neural network for the actuator model, which is combined with an analytical rigid-body simulator to generate training data for a quadruped robot \cite{hwangbo2019learning}, achieving thus direct transfer to the real quadruped robot.

Another key direction in combating the reality gap is to improve the generalization capability of the trained policies. One of the most commonly used techniques in this category is domain randomization, where a robust policy is trained with randomized simulator parameters for the robot \cite{peng2018sim, TanRSS18, akkaya2019solving, Siekmann-RSS-20, pinto2017robust}. By forcing the control policy to learn actions that can work for different simulation parameters during training, one can obtain a controller that is robust and generalizable to parameters outside the training range. To train an effective robust policy, researchers have explored numerous ways of introducing the randomization to the simulation model, such as adding adversarial perturbation \cite{pinto2017robust}, adding latency in the model \cite{TanRSS18}, and varying the dynamics parameters of the robot \cite{peng2018sim}. The design of these randomization schemes are largely inspired by the factors that contribute to the reality gap. For example, Tan \etal observed that latency and actuator modeling are two major sources of the reality gap and by including them in the randomization scheme they were able to achieve successful sim-to-real transfer for a quadruped robot \cite{TanRSS18}. On the other hand, kinematics-related parameters, such as the length of the robot's limb or the placement of the actuator, are rarely explored in domain randomization as they can usually be measured to high precision and be recreated in the simulation almost exactly. As a notable exception, the idea of including kinematic variations during training in simulation has been previously proposed in \cite{chen2018hardware}. Therein, the authors generate hardware-conditioned policies with both explicit and implicit representation. They show good robustness and transfer results on manipulation tasks. %, but focus mainly on transfer between robots of different hardware configuration. Furthermore, the considered fine-tuning approach requires a significant amount of rollouts in the target environment. 
In comparison, we demonstrate that randomizing kinematics-related parameters alone can in fact produce surprisingly strong performance compared to randomizing dynamics-related parameters, and present a more efficient adaptation scheme requiring only a small number of rollouts.

Another approach for improving the generalization capability of the trained control policies is to fine-tune them using data collected from the real robot \cite{yu2019sim, yu2019learning, peng2020learning, song2020rapidly, Chebotar19}. For example, Yu \etal proposed to learn a dynamics-conditioned policy in simulation and directly optimize the input dynamics parameters to the policy on the real hardware \cite{yu2019sim}, showing results on a real biped robot. Recently, Peng \etal \cite{peng2020learning} adopted a similar strategy and demonstrated that a real quadruped robot can learn from animal data by leveraging simulation and transfer learning \cite{peng2020learning}. Similar to prior work in domain randomization, these methods mostly focus on dynamics-related variations during the training of the policies and do not consider kinematics-related variations. In our work, we demonstrate that introducing kinematics-based variations in this class of methods also leads to improved sim-to-real performance. We further introduce an ensemble model-based approach that further improves the reliability of the framework proposed by Yu \etal \cite{yu2019sim}.

\section{Background and Selection of Domain Randomization Parameters}
We begin by presenting a few preliminaries useful for the problem under consideration.  The  robot task is formulated as a Markov Decision Process (MDP) $(\mathcal{S}, \mathcal{A}, f, r, p_0, \gamma)$, where $\mathcal{S}$ is the state space,  $\mathcal{A}$ is the action space, $f(\cdot)$ is the system dynamics, $r(\cdot)$ is the reward function, $p_0$ is the initial state distribution, and $\gamma$ is the discount factor. We use a policy gradient method, PPO~\cite{Schulman:2017}, to solve for a policy $\pi$ such that the accumulated reward is maximized:
\begin{equation}
    J(\pi) = \mathbb{E}_{\mathbf{s}_0, \mathbf{a}_0, \dots, \mathbf{s}_T} \sum_{t=0}^{T} \gamma^t r(\mathbf{s}_t, \mathbf{a}_t),
\end{equation}
where $\mathbf{s}_0 \sim p_0$, $\mathbf{a}_t \sim \pi(\mathbf{a}_t|\mathbf{s}_t)$, and $\mathbf{s}_{t+1}=f(\mathbf{s}_t, \mathbf{a}_t;\p)$. 
%\yifeng{I used f elsewhere so need to replace this one -- also it would be better if we include design/dynamics parameters into $f$ to show that we solve a family of MDPs?}
Here, $\p \in \mathcal{P}$ is a set of parameters pertaining to the robot that we can vary during training in simulation. %\yannis{<-- this last sentence needs to be modified. $\p$ is not just design anymore...} \wenhao{modified to be all the simulation parameters, is that correct?} 
%These include, but are not limited to, the robot's design, for example.
We divide these parameters $\p$ in two categories, kinematic parameters ($\p_k$) and dynamic parameters ($\p_d$). Kinematic parameters include those required for forward kinematic computation, assuming the robot can be represented as an articulated rigid body system. Specifically, $\p_k$ includes every coefficient in the kinematic transformation chain from the robot frame to the local frame of every link. For example, the coefficients required to compute the transformation of the ``foot frame'' includes the link lengths of the upper leg and the lower leg, the orientation of the hip, knee and ankle joints, and the location of the hip joint relative to the torso. The definition of dynamic parameters is simply $\{\p_d\} = \{\p\} \setminus\{ \p_k\}$.

We further define three types of sim-to-real gaps frequently encountered in robotic applications (Concrete examples of sim-to-real gaps are shown in Section \ref{Sec:exp}):\\
%\begin{itemize}
 %   \item 
 \textbf{Kinematic Gap}: One or more $\p_k$ have different values between source and target environments. For example, the length or joint orientation of the robot legs are different.\\
\textbf{Dynamic Gap}: One or more $\p_d$ have different values between the source and target environments. For example, the mass distribution of the robot or the actuator modeling are different.\\
\textbf{Environment Gap}: Other parameters outside of $\p$ in the simulator are different between the source and target environments. For example, the surfaces the robot walks on are made of different materials.
%\end{itemize}

We randomize a selected set of $\p_k$ and $\p_d$ and train two types of policies, $\pi_k$ and $\pi_d$, using PPO \cite{Schulman:2017}. These two polices are then tested on three types of  sim-to-real gaps. Note that we intentionally model these gaps using parameters different from those being randomized. The detailed results are shown in Section \ref{Sec:exp}, but here we highlight a few key results. First, $\pi_k$ performs well when transferred over a kinematic gap and $\pi_d$ performs well when transferred over a dynamic gap. Second, $\pi_k$ outperforms $\pi_d$ significantly when transferred over the opposite gap, as well as the environment gap.  We hypothesize that randomizing kinematic parameters results in wider exploration of the state space because it has a more global impact on the system dynamics. A small change in $\p_k$ will affect the kinematic transformation in the Jacobian matrix, which in turn affects the mass matrix, Coriolis force, gravitational force, and every Cartesian position-dependent applied force, such as contacts. In contrast (and perhaps counter-intuitively), randomizing dynamic parameters affects only specific aspects of the dynamic system. For example, inertial properties only take effect when the joint acceleration is high, and friction coefficients only matter when particular contact states occur.

\section{Domain Adaptation using Kinematic Parameters}

Domain Adaptation techniques refer to a class of transfer methods that train a family of policies (universal policies, UP), conditioned on varying explicit physical parameters (e.g \cite{yu2018policy}), or implicit latent parameters (e.g. \cite{yu2019sim, peng2020learning}). During simulation-based training, the environment dynamics change with the conditioned parameters so that we obtain different strategies good for different dynamics, parameterized by different conditioned input. When deploying the universal policy to the target environment, the conditioned input (purely virtual, as the geometry of the physical robot is fixed) can be quickly searched to find a well-performing strategy in the target domain with only a few rollouts, using sample efficient optimizers such as Bayesian Optimization \cite{mockus2012bayesian}. The key insight of these methods is that the conditioning parameters that lead to good performance do not need to have physical correspondence (for latent space methods they have no physical meaning at all), the only relevant aspect being that learned strategies are diverse enough so that the likelihood of good adaptation is high.

As we shall see in results of Section \ref{Sec:exp}, an adapted UP can potentially outperform a policy trained with kinematic domain randomization. However, we also observed that the transfer ability of a policy in the target environments varies across training seeds, despite all seeds producing policies with virtually identical training environment performance. This seems to imply that policy transfer performance could be further improved if we were able to perform UP-based policy adaptation with a policy ensemble rather than one single UP as in previous works.  However, the problem of allocating limited target domain rollouts to multiple polices during adaptation poses a challenge. %As such, we propose Multi-Policy Bayesian Optimization (MPBO), an algorithm developed to intelligently allocate sample budget by prioritizing sampling from the most promising policies.
% As a result, we need to diversify our approach by training policies for at least a few seeds; this is not a particular restriction as training is done in simulation and is thus inexpensive. Tuning these policies however requires rollouts, which are limited in the target environment. 
We explore how to prioritize policy and parameter value sampling on a limited budget of rollouts in what follows.

\subsection{Multi-Policy Bayesian Optimization}
% \yifeng{todo: explain evaluation function, which takes p to the policy j and run rollouts in the target env to get mean reward; explain adjusted reward with minus std} 
Given an ensemble of UPs trained with different seeds, we wish to bias target domain sampling towards the current most promising UP, but without over-exploitation. We propose Multi-Policy Bayesian Optimization (MPBO), a hybrid procedure based on a combination of Bayesian optimization (BO) \cite{mockus2012bayesian} and the Upper-Confidence-Bound Action Selection (UCBAS) algorithm \cite{sutton2018reinforcement} from the field of multi-armed bandit problems. With a reward evaluation function $E$ which takes in a conditioning parameter sample and the universal policy as input and returns accumulated reward, BO progressively builds a Gaussian process (GP) regressor \cite{williams2006gaussian} of input vs. obtained reward for that policy based on the history of samples. This function is then used to create an \textit{acquisition function}, whose role is to suggest new sample points \cite{mockus2012bayesian}. In MPBO, we accordingly create an ensemble of GP, one for each UP. The acquisition function returns a new sample point suggestion (i.e., a new parameter input value) for each policy, along with a metric called \textit{expected improvement} ($e$), which indicates how useful the new sample is expected to be for the optimization. We only sample from the most promising policy/parameter value at the current iteration, rather than sampling from all of them. The process of selecting which policy to sample is similar to a multi-armed bandit problem: we wish to prioritize the sampling of policies that seem promising overall, but also explore policies that have received less attention. Specifically, we adopt the technique ``Upper-Confidence-Bound Action Selection'' which suggests sampling the option with the highest following value:
\begin{equation}
\mathrm{u}_j = \bar{R}_j + c \sqrt{ \frac{\ln(t)}{N_j}}
\end{equation}
where $\bar{R}_j$ is the average reward (over all samples) for that option so far, $t$ is the algorithm iteration number (starting from 1), $N_j$ is the number of times this option has been selected before, $j$ is the option index, and $c$ is a constant that balances exploration vs. exploitation. In our context, $\bar{R}_j$ would be the mean reward that policy $j$ has produced for all its sampled parameters values so far and $N_j$ the number policy $j$ has been selected for sampling. Combining this metric with  \textit{expected improvement} ($e$), the final selection criterion is sampling the policy/parameter value with the highest product $u \cdot e$. After the budget of rollouts has been exhausted, MPBO returns the best policy, best parameter input, and expected reward. The MPBO procedure is summarized in Alg.~\ref{alg:MPBO}.

\begin{algorithm}[t]
\SetAlgoLined

\SetKwInput{KwInput}{Require}
\KwInput{Ensemble of $M$ Universal Policies \\ 
GP Ensemble: $\{\mathcal{G}\}_{j=1}^M$ \\
Exploration factor: $c$\\
Parameter Buffers of each policy: $\{\vc{P}\}_{j=1}^M \leftarrow \{\{ \}\}_{j=1}^M $ \\
Reward Buffers of each policy: $\{\vc{R}\}_{j=1}^M \leftarrow \{\{ \}\}_{j=1}^M $ \\
Target env. reward function: $E^j(\vc{p}):=E(\vc{p};\pi_j)$, $j = 1,\ldots M$
}
\vspace{5pt}
// initialize the GPs: \;

\For{$j = 1,\cdots,M$}{
    $\{\vc{p}^j\} \leftarrow \textrm{random()}$ \;
    
    $\vc{P}^j, \vc{R}^j \leftarrow \textrm{store}(\{\vc{p}^j\},\{E^j(\vc{p}^j) \})$ \;
    
    $\mathcal{G}^j \leftarrow \textrm{fit}(\vc{P}^j, \vc{R}^j)$ \;
}
\vspace{5pt}
// $p$ search guided by MPBO \;

$t \leftarrow 1$ \;

\While{\textrm{budget not used up}}{
    \For{$j = 1,\ldots,M$}{
        $\vc{p}^j, e^j \leftarrow \textrm{sampleNext}(\vc{p} | \mathcal{G}^j )$ \;
        
        $u^j \leftarrow \textrm{mean}(\vc{R}^j) + c\sqrt{\ln(t)/\mathrm{len}(\mathbf{R}_j)}$ \;
    }
    $j^* \leftarrow \arg\max_j (u^j \cdot e^j$) \;
    
    $\vc{P}^{j^*}, \vc{R}^{j^*} \leftarrow \textrm{store}(\vc{p}^{j^*}, E^{j^*}(\vc{p}^{j^*}) )$ \;
    
    $\mathcal{G}^{j^*} \leftarrow \textrm{fit}(\vc{P}^{j^*}, \vc{R}^{j^*})$ \;
    \vspace{5pt}
    // other GPs not updated \;
    
    $t \leftarrow t + 1$ \;
}

\caption{Multi-Policy Bayesian Optimization (MPBO)}
\label{alg:MPBO}
\end{algorithm}

\section{Experiments}\label{Sec:exp}

In this section, we design a set of experiments to answer the following questions: A. Do both kinematics and dynamics domain randomization transfer well when the actual domain discrepancy is related by type to their randomized aspects? B. Does kinematics domain randomization achieve better transfer performance on unmodelled discrepancies compared to the typical dynamics domain randomization? C. Can a kinematics domain UP optimized with MPBO further improve the transfer performance?

\subsection{Experiment setup}
\label{ssec:exp_setup}

\begin{figure}[t]
\centering
\subfigure[]{
  \includegraphics[width=0.48\linewidth]{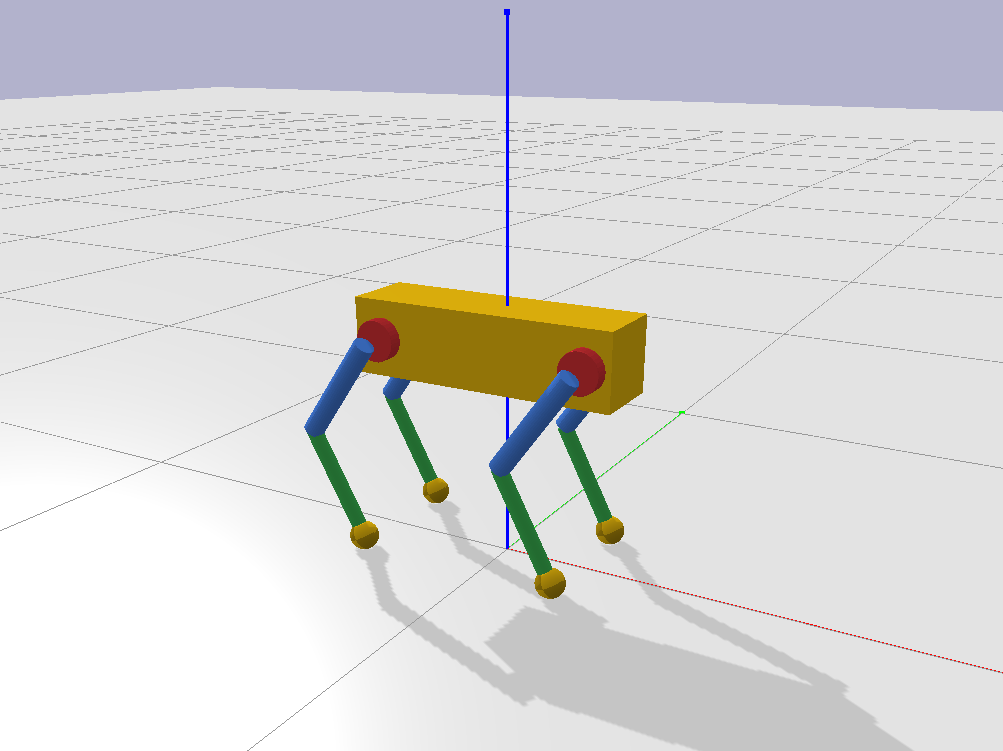}}
\hspace*{-2mm}
\subfigure[]{
  \includegraphics[width=0.48\linewidth]{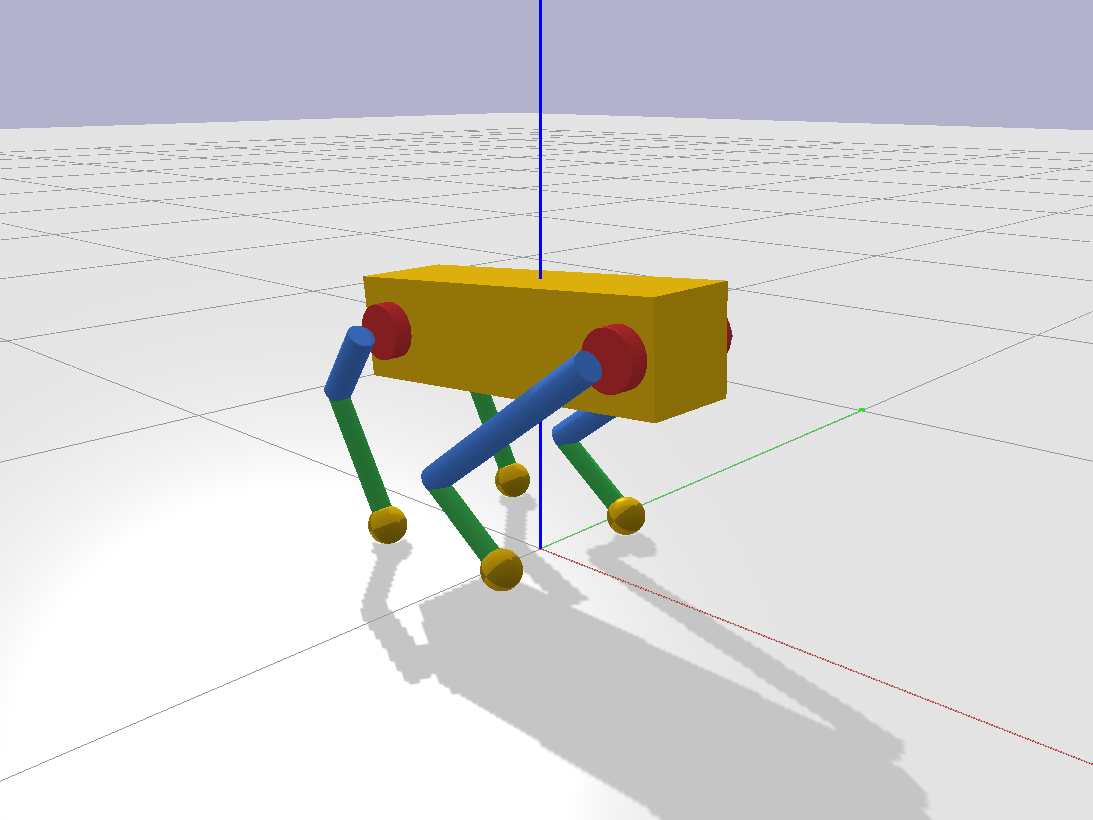}}
%\hspace*{-5mm}
 \vspace{-4mm}
\caption{The quadruped locomotion environment. (a) The nominal design used in target environments, a quadruped robot modelled from the Unitree Laikago \cite{laikago2018}. (b) a randomly sampled virtual kinematics design used during training only.} \label{fig:laikago_robot}
\end{figure}

We use a 18-DoF quadruped robot modelled from the Unitree Laikago \cite{laikago2018} as our experiment platform. The robot is simulated in PyBullet \cite{coumans2017pybullet}. Fig.~\ref{fig:laikago_robot}(a) shows its real kinematics structure. To introduce virtual kinematics variations during simulation-based training, we consider the length scale of the two links of the front leg pair as well as the back leg pair, leading to a total of four parameters: $\p = [p_1, p_2, p_3, p_4]$, where $p_1$ and $p_2$ are the scales of the upper and lower link of the front legs, and $p_3$ and $p_4$ are the scales of the upper and lower link of the back legs. These parameters are drawn uniformly in $[50, 150]\%$ both in Kinematics Randomization and Kinematics UP, that is, ranging from a 50\% decrease to a 50\% increase in link length. A randomly sampled virtual design is shown in Fig.~\ref{fig:laikago_robot}(b).
All generated policies consist of a 3-layer feed-forward neural network where the  input is the robot observation along with the kinematics parameters in the case of kinematics-UPs. The kinematics parameters are constant throughout each rollout, in a manner similar to goal-conditioned polices. We apply position control and the policy output corresponds to the change in desired joint positions, where the position errors are used to compute the required torque with a PD-control scheme. The observation vector consists of the joint states along with root global orientation, root height and root linear velocities. Control frequency is set to 50Hz. The reward function is given by 
$$r(\mathbf{s},\mathbf{a}) = 4.5+4v_x -0.1\|\mathbf{a}\|^2-0.5\|\mathbf{j}\|_0-0.001\|\ddot{\mathbf{q}}\|$$
to encourage walking forward with smooth joint motion and low energy, where $v_x$ is the root forward velocity, $||\mathbf{j}||_0$ is the number of joints at joint limit, and $\ddot{\mathbf{q}}$ is the acceleration of the robot's generalized coordinates.

\subsection{Target Environments}
\begin{figure}[t]
\centering
\subfigure[]{
  \includegraphics[width=0.48\linewidth]{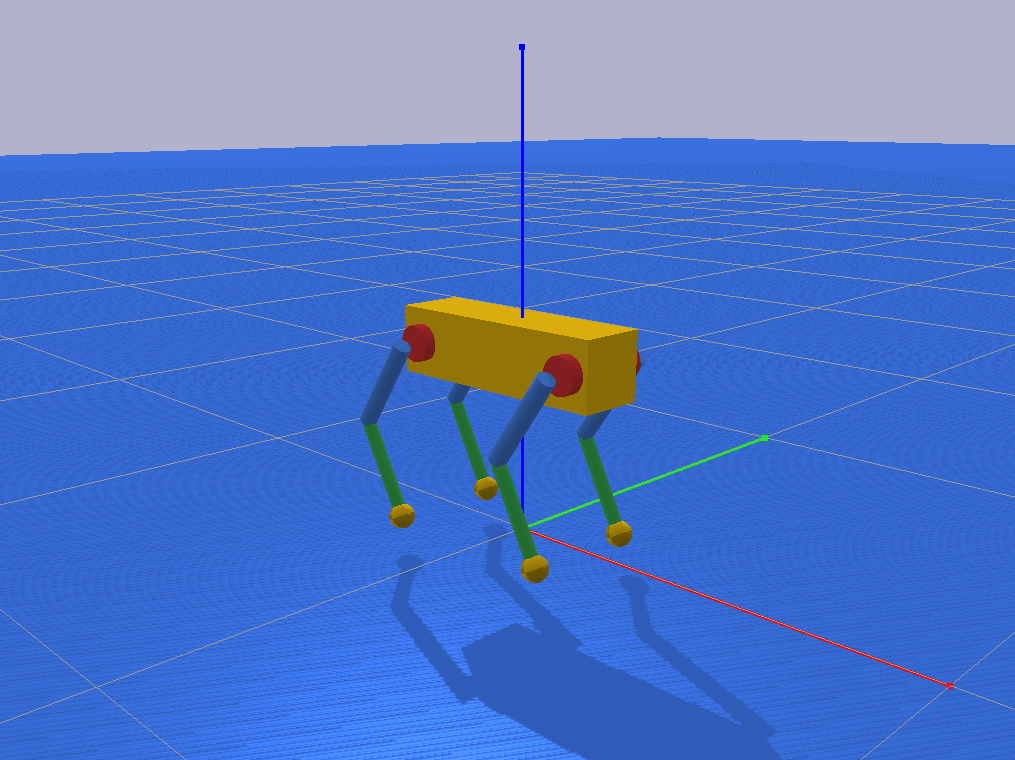}}
\hspace*{-2mm}
\subfigure[]{
  \includegraphics[width=0.48\linewidth]{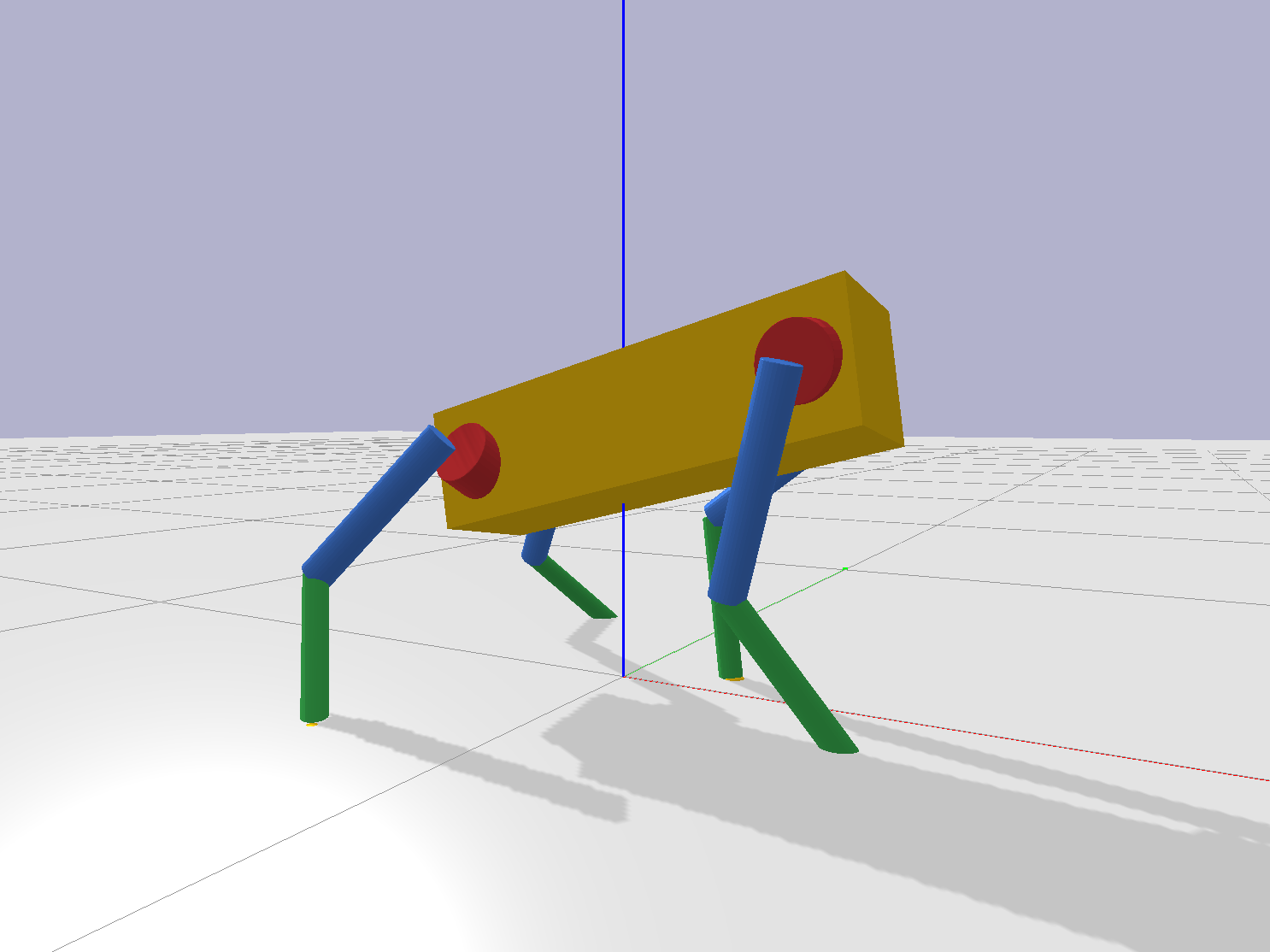}}
%\hspace*{-5mm}
 \vspace{-4mm}
\caption{Target environments for the quadruped. (a) Deformable floor: the robot is placed on an elastic platform in the form of a deformable cube. (b) Soft floor: the robot legs sink in. } \label{fig:test_env}
\end{figure}

We consider the following types of gaps between training and target settings: (a) dynamics gaps originating from modeling errors inherent to robot dynamics, (b) kinematic gaps representing discrepancies in the robot kinematics, and (c) environment gaps where unmodelled effects are introduced to the surrounding environment that the robot is moving within. To this end, we evaluate the ability of policies to transfer to unseen environments by introducing five target environment variants:\\
%\begin{enumerate}
(1) \textbf{Dynamics -- Low Power}: The maximum allowed torque in the front left leg is reduced by a half, to mimic broken motors. 
(2) \textbf{Dynamics -- Back-EMF}: Increased joint angular velocity reduces the ability to apply torque to mimic Back-EMF motor forces.
(3) \textbf{Kinematics -- Joint Orientation}: All set points (zero angle) of the robot joints are perturbed by different constants. Note that this is not in the list of kinematics randomization parameters, which only includes link lengths.
(4) \textbf{Environment -- Deform}: Here, the fixed and solid floor is replaced by a large deformable cube, see Fig.~\ref{fig:test_env}(a). Note that deformable bodies are simulated in PyBullet independently using FEM and parameters controlling its properties are thus not related to randomized rigid-body parameters such as restitution.
(5) \textbf{Environment -- Soft}: In this target environment the floor is soft and the robot legs sink in, in a similar manner to a muddy terrain, Fig.~\ref{fig:test_env}(b).\\
%\end{enumerate}
These gaps are made challenging enough so that a policy trained to convergence without randomization will perform poorly when in them (Table \ref{tab:results} first row). More importantly, they emphasize unmodelled effects not covered by the selection randomized parameters, which mimics the realistic challenge that there are always aspects on the real hardware which we cannot randomize well.

\subsection{Baselines}
\label{ssec:baselines}
To evaluate the performance of kinematics randomization used to create robust policies for zero-shot transfer, we utilized the following baselines: (A) \textbf{No Domain Randomization (No DR)}: a conventional PPO policy trained without domain randomization.
%\item 
(B) \textbf{Dynamic parameter randomization (Dynamic DR)}: we train a robust policy with randomized robot dynamics, using the same randomization settings as Peng \etal which have been demonstrated on a real robot \cite{peng2020learning}.\\
%\end{itemize}
For evaluating kinematic parameter UPs and MPBO, we introduce the following baselines that perform adaptation in the target environment: (C) \textbf{No DR with fine-tuning}: we fine-tune the No DR policy with additional PPO steps in the target environment, using approximately 20 rollouts for each seed.
(D) \textbf{Dyn-UP + 10 rollouts}: UPs conditioned on dynamic parameters during training. We follow the latent dimension scheme by Yu \etal \cite{yu2019sim}, where the high dimensional dynamics parameters (52 in our case) is mapped to a low dimensional latent space (4 in our experiments to match Kin-UP) and the adaptation is performed directly in the latent space. Three policies are generated for three seeds, and each UP is adapted in its latent dimension using regular BO for 10 rollouts, for a budget total of 30 rollouts.
(E) \textbf{Kin-UP, pre-adapt. / Kin-UP + 10 rollouts}: UPs conditioned on kinematic parameters during training. Three policies are generated for three seeds. For the pre-adapt. version, the best performance for nominal parameter input values among the three UPs is reported (no adaptation), whereas for the 10 rollouts version, each of the three UPs is adapted for 10 rollouts, for a total budget of 30 rollouts.\\
%\end{itemize}
The proposed adaptation scheme, Kin-UP + MPBO consists of the aforementioned Kin-UPs for three seeds, with adaptation of a total budget of 30 rollouts spent \textit{unequally} among them, in accordance with MPBO. To provide a fair comparison to our proposed MPBO algorithm where three policies trained from different random seeds are utilized to adapt to one final policy selection, we trained three policies for all baseline methods, and for each target environment, we report the return corresponding to the \textit{best} performing policy among the three. Final performance evaluations are averaged over 15 rollouts to provide a more accurate performance estimate. 

% \usepackage{multirow}
% \usepackage{multirow}
\begin{table*}[!ht]
\centering
\begin{tabular}{|c|c|c|c|c|c|c|c|c|}
\hline
\multicolumn{2}{|c|}{\multirow{2}{*}{Method}} &
  \multirow{2}{*}{No gap} &
  \multicolumn{2}{c|}{Dynamic Gap} &
  Kinematic Gap &
  \multicolumn{2}{c|}{Environment Gap} &
  \multirow{2}{*}{\begin{tabular}[c]{@{}c@{}}Gap \\ Average\end{tabular}} \\ \cline{4-8}
\multicolumn{2}{|c|}{} &
   &
  Low power &
  EMF &
  Joint orientation &
  Deform &
  Soft &
   \\ \hline
\multirow{3}{*}{\begin{tabular}[c]{@{}c@{}}Zero-\\ Shot\end{tabular}} &
  No DR &
  9458 &
  597 &
  4980 &
  79 &
  1636 &
  1368 &
  1732 \\ \cline{2-9} 
 &
  Dynamic DR &
  8346 &
  \textbf{8010} &
  \textbf{5635} &
  220 &
  1122 &
  508 &
  3099 \\ \cline{2-9} 
 &
  \begin{tabular}[c]{@{}c@{}}Kinematic DR\\ (ours)\end{tabular} &
  9141 &
  7463 &
  5112 &
  \textbf{5401} &
  \textbf{5795} &
  \textbf{7305} &
  \textbf{6215} \\ \hline\hline
\multirow{5}{*}{Adapt.} &
No DR + Fine-tuning &
  - &
  600 &
  5267 &
  80 &
  1386 &
  4334 &
  2334 \\ \cline{2-9} 
 &
  Dyn-UP + 10 rollouts per policy &
  - &
  \textbf{7861} &
  4721 &
  223 &
  1407 &
  1538 &
  3150 \\ \cline{2-9} 
 &
  Kin-UP pre-adapt &
  - &
  2262 &
  4967 &
  657 &
  2657 &
  477 &
  2204 \\ \cline{2-9} 
 &
  Kin-UP + 10 rollouts per policy &
  - &
  7249 &
  5564 &
  7040 &
  3246 &
  6927 &
  6005 \\ \cline{2-9} 
 &
  \begin{tabular}[c]{@{}c@{}}Kin-UP+MPBO\\ 30 rollouts (ours)\end{tabular} &
  - &
  7768 &
  \textbf{5572} &
  \textbf{7277} &
  \textbf{8250} &
  \textbf{7375} &
  \textbf{7248} \\ \hline
\end{tabular}
\caption{Results for our experiments. For each method, three policies are generated using different seeds. %The reported results are a performance average over 15 rollouts for the best policy (MPBO indicates the specific policy to be used itself).
}
\vspace{-2mm}
\label{tab:results}
\end{table*}

\subsection{Results for Kinematic Domain Randomization}
We first present our results for training a kinematic domain randomization (kinematic DR) policy and compare them to No DR and dynamics DR as described in Section \ref{ssec:baselines}. The results can be found in the top part of Table~\ref{tab:results}. Unsurprisingly, training without any randomization (No DR) performs the worst in the target setting despite the high reward obtained in the training environment. This validates that our target environments are substantially different from the training setting. Dynamics DR, on the other hand, achieves decent performance for the target environments where the discrepancies lie mainly in the robot dynamics. This is likely because the policy has been trained with a variety of robot dynamics and is thus robust to dynamics-related variations even though our target environment consists of dynamics gaps that were not presented during training. However, the results also indicate that when we apply dynamics DR to a target setting where the kinematics and the environment parameters are varied, it is not able to achieve successful transfer. In contrast, our proposed kinematic DR approach is able to reach high performance for all the target scenarios, and furthermore significantly outperforms the baseline methods in target settings with kinematic and environment gaps.

\subsection{Results for Multi-Policy Bayesian Optimization (MPBO)}

Having seen the promising performance of Kinematic DR, a natural question to ask is: can we further improve the transfer performance of the algorithm by allowing policies to adapt in the target environment? In the bottom part of Table \ref{tab:results}, we present our experiment results for our proposed MPBO algorithm applied to a universal policy (UP) conditioned on kinematics parameters of the robot. To illustrate the effect of the adaptation process, along with MPBO, we also report the performance of the trained UP with the nominal design as input, i.e. no input adaptation is performed. As this is analogous to training a single policy at the nominal design, the performance is similarly poor to No DR. However, as shown in our results, performing adaptation using MPBO for $30$ rollouts %(15 MPBO iterations with 2 rollouts per reward function $E$ evaluation)
in the target environment outperforms regular BO for 10 rollouts on each one of them individually. Comparing Kin-UP + MPBO to a baseline method where we fine-tune the No DR policy with an at least equal number of rollouts as well as the three Dyn-UPs updated by 10 rollouts each (Section \ref{ssec:baselines}),  we observe that fine-tuning the No DR policy with PPO typically achieves little improvement, and Dyn-UP performs slightly better only in one environment and trails behind in the rest.

\section{Conclusions and Discussion}
The combined results of Table~\ref{tab:results} suggest that introducing variations in \textit{kinematics} during training in simulation can benefit policy transfer to novel settings that are not seen during training. Perhaps surprisingly, this approach substantially outperforms the commonly used dynamics domain randomization in a variety of novel target environment settings. One hypothesis behind this unintuitive observation is that varying kinematic parameters during training leads to more global changes in the dynamic equations which subsequently leads to the wider area of the state space that is visited. Further investigation is needed to validate this hypothesis. 

% \yifeng{Did we forget to explain the 2D slices?}
% \yannis{They are in Conclusions and discussion. Would it be better to put them in the results section? Though they are mostly supportive evidence to arguments we make in discussion...} \wenhao{I think it'll be less confusing once the paper is trimmed an the figure will lie on the same page as the text discussing it.}
\begin{figure}[t]
\centering
%\subfigure[]{
%  \includegraphics[width=0.48\linewidth]{figures/2Dsoft.png}}
%\hspace*{-2mm}
%\subfigure[]{
%  \includegraphics[width=0.48\linewidth]{figures/2Dsoft.png}}
%\hspace*{-2mm}
\subfigure[]{
  \includegraphics[width=0.46\linewidth]{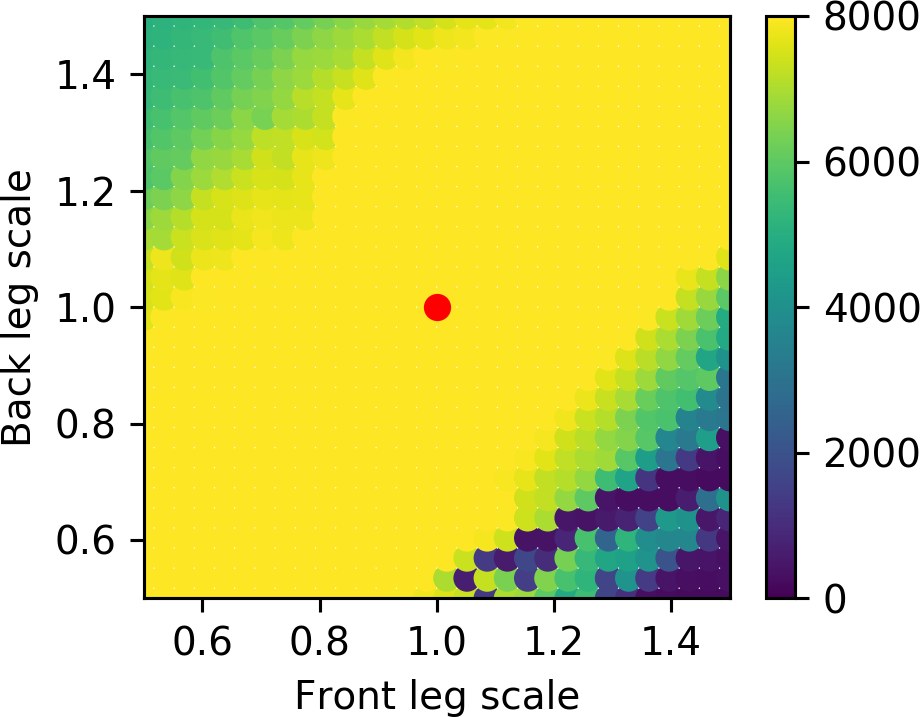}}
\hspace*{-2mm}
\subfigure[]{
  \includegraphics[width=0.46\linewidth]{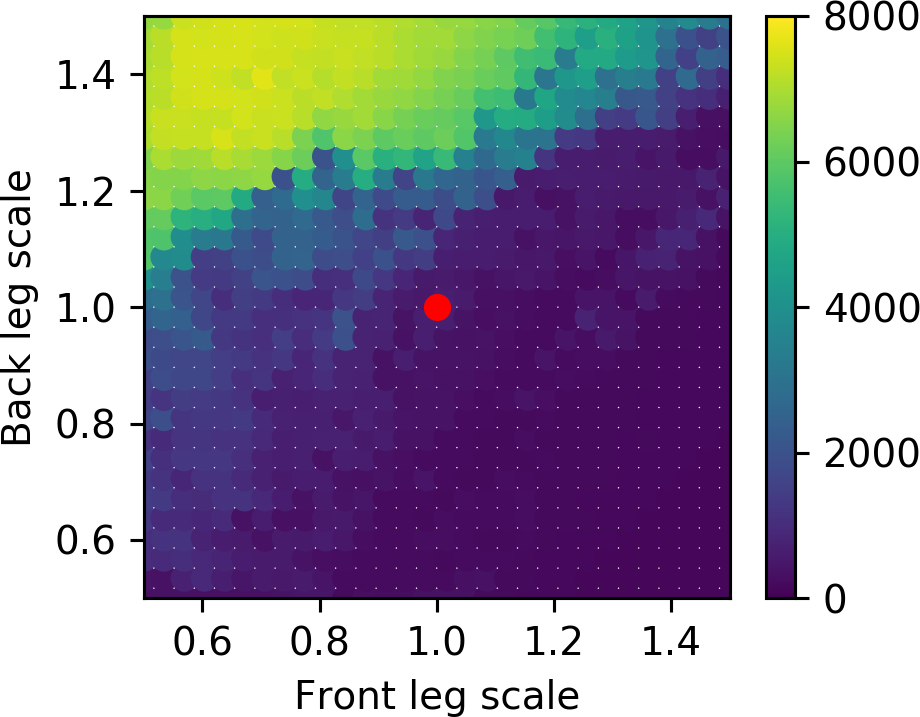}}
\subfigure[]{
  \includegraphics[width=0.46\linewidth]{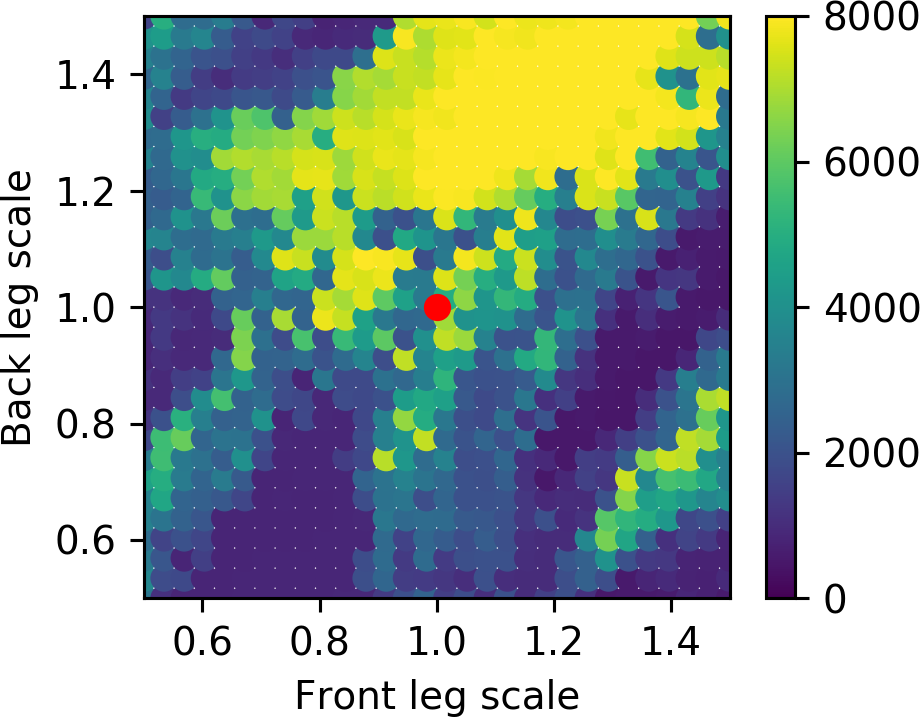}}
\hspace*{-2mm}
\subfigure[]{
  \includegraphics[width=0.46\linewidth]{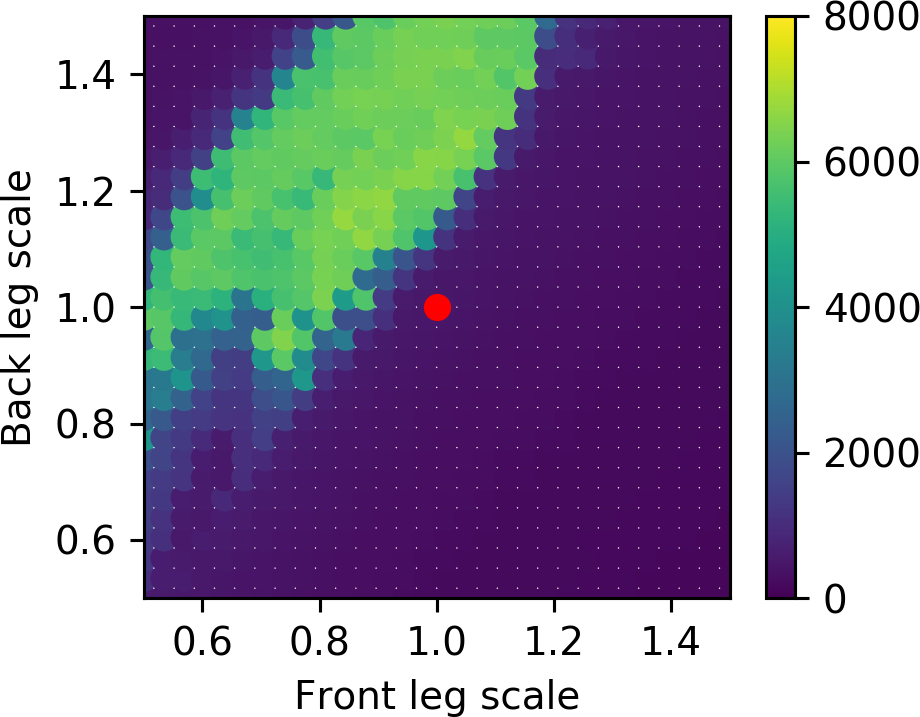}}
\caption{2D slices of the reward landscape at different targeting environments as function of the parameter input to the policy. The actual kinematic parameters are set to the nominal value (i.e., $\mathbf{1}$), shown as the red dot. The 4D parameters are tied in groups of two, i.e., $[x, x, y, y]$ and evaluated over a $(x,y)$-grid. (a) no gap (training environment), 
(b) joint orientation, (c) deformable floor, (d) soft floor.%, (e) low leg power, (f) EMF.
}
\vspace{-15pt}
\label{fig:2d}
\end{figure}

%Gaps encountered in the testing environment also effectively modify the visited state space. Thus, it is plausible that there might be some overlap between changes in the visited state space among the two. A policy which has been exposed to such changes during training may perform better than one which has never done so It also seems reasonable that variations in kinematics can diversify the visited state space to a greater extent than variations in dynamic parameters, and that may be the reason why the kinematics-based approach outperforms dynamic parameter randomization. 

Another key result in our experiment is that with limited amount of data in the target environment, we can significantly improve the transfer performance using an improved version of strategy optimization \cite{yu2018policy}. By exploiting inexpensive computation in simulation, we can turn the variance due to random seeds in policy optimization to our advantage and extensively explore multiple universal policies at different local minima. We then rely on our MPBO to find the most effective strategy among multiple families of policies. 

%We speculate that our hypothesis regarding the superior performance of kinematics DR may also justify the benefit of MPBO: 1) a gap causes a change in the visited state space, 2) the kinematic DR policy $\pi_k$ has not encountered that state space area, 3) through fine tuning, the policy is asked to act according to an imaginary design that would produce good results if the policy visited that area of the state space during training. 

%Some partial evidence for the latter hypothesis
%\yannis{this needs to blend with the preceding text} \wenhao{Changed the first sentence. It seems the hypotehsis this visualization is intended of is removed? }

To investigate the mechanism behind the success of MPBO, we depict the reward landscape of a kinematic UP over the kinematic parameters. Specifically, we obtain a 2D slice of the 4D landscape by tying parameters in groups of two, i.e., by varying $[x, x, y, y]$. We then evaluated the Kin-UPs on such a grid of $(x,y)$, with each sampled parameter input being tested for 15 rollouts. The results for the training environment and three target environments (joint orientation, deformable, and soft floor) are depicted in Fig.~\ref{fig:2d}. For the training environment, maximum performance is obtained at an area near the actual kinematic parameters of the robot, as expected. However, for target environments the default kinematic parameter lies in a low reward region and performance can be greatly increased by modifying the kinematic parameter input to the policies. Furthermore, the reward landscape shows some underlying structure; thus, the performance increases due to MPBO seen in Table~\ref{tab:results} are not merely attributed to randomness, but are rather the outcome of a diversified policy, whose tuning knobs exert enough influence over the policy to successfully bridge the gaps.  

This work opens a few interesting future directions. While our experiments show strong signals that randomizing kinematic parameters is beneficial, it requires broader investigation on other types of robot tasks, such as manipulation, to validate whether our results can be generalized. The hypothesis we put forth is based on our intuition of multi-body dynamic systems. A formal proof or empirical evidence to support our hypothesis is also an important potential future direction.% we would like to investigate.

%\newpage

\addtolength{\textheight}{-12cm}   % This command serves to balance the column lengths

%%%%%%%%%%%%%%%%%%%%%%%%%%%%%%%%%%%%%%%%%%%%%%%%%%%%%%%%%%%%%%%%%%%%%%%%%%%%%%%%

%%%%%%%%%%%%%%%%%%%%%%%%%%%%%%%%%%%%%%%%%%%%%%%%%%%%%%%%%%%%%%%%%%%%%%%%%%%%%%%%

%%%%%%%%%%%%%%%%%%%%%%%%%%%%%%%%%%%%%%%%%%%%%%%%%%%%%%%%%%%%%%%%%%%%%%%%%%%%%%%%
%\section*{APPENDIX}


%\section*{ACKNOWLEDGMENT}

%%%%%%%%%%%%%%%%%%%%%%%%%%%%%%%%%%%%%%%%%%%%%%%%%%%%%%%%%%%%%%%%%%%%%%%%%%%%%%%%

%References are important to the reader; therefore, each citation must be complete and correct. If at all possible, references should be commonly available publications.

\bibliographystyle{IEEEtran}
\bibliography{IEEEabrv.bib,references}

\end{document}